\documentclass[10pt,twocolumn,letterpaper]{article}

\usepackage[pagenumbers]{cvpr} 

\usepackage{algorithm}
\usepackage{algorithmic}
\usepackage{newfloat}
\usepackage{listings}
\usepackage{subcaption}
\usepackage{amsthm,amsmath,amssymb}
\usepackage{mathrsfs}
\usepackage{multirow,booktabs}
\usepackage{graphicx}

\definecolor{cvprblue}{rgb}{0.21,0.49,0.74}
\usepackage[pagebackref,breaklinks,colorlinks,allcolors=cvprblue]{hyperref}

\def\confName{CVPR}
\def\confYear{2026}

\title{Association and Consolidation: Evolutionary Memory-Enhanced Incremental Multi-View Clustering}

\author{Zisen Kong$^{1,2}$, Bo Zhong$^{1,2}$, Pengyuan Li$^{1,2}$, Dongxia Chang$^{1,2,*}$, Yiming Wang$^{3}$, Yongyong Chen$^{4}$ \\
$^{1}$Institute of Information Science, Beijing Jiaotong University, Beijing, China\\
$^{2}$Visual Intelligence + X International Cooperation Joint Laboratory of MOE, Beijing, China\\
$^{3}$School of Computer Science, Nanjing University of Posts and Telecommunications, Nanjing, China\\
$^{4}$School of Computer Science and Technology, Harbin Institute of Technology, Shenzhen, China\\
{\tt\small \{zskong@bjtu.edu.cn,dxchang@bjtu.edu.cn,ymwang@njupt.edu.cn,YongyongChen.cn@gmail.com\}}}

\begin{document}
\maketitle
\begin{abstract}
Incremental multi-view clustering aims to achieve stable clustering results while addressing the stability-plasticity dilemma (SPD) in view-incremental scenarios. The core challenge is that the model must have enough plasticity to quickly adapt to new data, while maintaining sufficient stability to consolidate long-term knowledge. To address this challenge, we propose a novel Evolutionary Memory-Enhanced Incremental Multi-View Clustering (EMIMC), inspired by the memory regulation mechanisms of the human brain. Specifically, we design a rapid association module to establish connections between new and historical views, thereby ensuring the plasticity required for learning new knowledge. Second, a cognitive forgetting module with a decay mechanism is introduced. By dynamically adjusting the contribution of the historical view to optimize knowledge integration. Finally, we propose a knowledge consolidation module to progressively refine short-term knowledge into stable long-term memory using temporal tensors, thereby ensuring model stability. By integrating these modules, EMIMC achieves strong knowledge retention capabilities in scenarios with growing views. Extensive experiments demonstrate that EMIMC exhibits remarkable advantages over existing state-of-the-art methods.
\end{abstract}


\section{Introduction}
\label{sec:intro}

Recent advancements in multimedia technology have significantly transformed data analysis paradigms, where Multi-View Clustering (MVC) has demonstrated notable advantages in unsupervised scenarios by integrating heterogeneous data sources \cite{AwMVC,du2025openviewer,wen2023,fu2024subspace}. Traditional MVC methods have achieved significant success in processing static, complete datasets, as they require a complete set of views to achieve effective knowledge fusion \cite{Du2024UMCGL}. However, data is not static but evolves dynamically over time in many real-world applications. For example, medical monitoring may extend from clinical measurements to imaging scan data, and multi-view data facilitates tracking of patient progress. This view-incremental property renders traditional MVC methods inefficient or even infeasible, as retraining the model from scratch with each new view fails to leverage previously learned knowledge. Consequently, Incremental Multi-View Clustering (IMVC) has become a critical research direction \cite{chen2025multi,wang2025adaptcmvc,feng2025incremental,CMVC}, which aims to update clustering results efficiently as new views arrive. However, its core challenge is to resolve a fundamental contradiction: the Stability-Plasticity Dilemma (SPD) \cite{wu2021striking, kim2020stability}. On the one hand, the model must possess sufficient plasticity to enable rapid learning and adaptation to new patterns and knowledge contained in new data. If the model is too rigid, it will be unable to capture dynamic changes in the data, leading to a rapid decline in performance. On the other hand, the model must also maintain sufficient stability to consolidate long-term accumulated knowledge and avoid catastrophic forgetting when learning new information. Therefore, an ideal IMVC model must achieve a balance between these two aspects.

Recent incremental multi-view clustering (IMVC) methods have shown significant progress. For example, Wan et al. \cite{CMVC} update a consensus partition matrix by transferring knowledge from historical views, enabling incremental learning. Yan et al. \cite{yan2024live} further introduced a category memory library to store learned categories, while Qu et al. \cite{LAIMVC} proposed a lightweight framework that retains and jointly optimizes the consensus anchor graph of previous views. Despite existing IMVC methods having made valuable explorations, they often address only specific aspects of SPD. Some prioritize plasticity, rapidly integrating new data at the risk of catastrophic forgetting, while others emphasize stability through regularization, potentially hindering adaptation to new views. A key limitation is that existing methods fail to balance old knowledge retention and new knowledge integration, which causes performance bottlenecks as views continuously grow. To address this dilemma, we drew inspiration from the memory system of the human brain. Cognitive science research indicates that the collaborative memory system between the hippocampus and the prefrontal cortex serves as a paradigm for balancing plasticity and stability. The hippocampus can quickly encode new episodic memories and is highly plastic \cite{arani2022learning}. In contrast, the prefrontal cortex integrates and consolidates these temporary memories into highly stable long-term knowledge through a slow process. This complementary mechanism enables humans to continue learning without easily forgetting the past. Furthermore, knowledge acquisition is not merely information accumulation. The brain also employs a complex active adaptive forgetting mechanism to discard outdated information and maintain cognitive efficiency \cite{anderson2021active}. This forgetting is not a passive process of memory decay but an active process of neural remodeling that occurs immediately after learning behavior is completed \cite{hardt2013decay}. 

Inspired by these cognitive principles, we propose an Evolutionary Memory-Enhanced Incremental Multi-View Clustering model (EMIMC) designed to address the SPD. We should clarify that EMIMC does not aim for biologically accurate simulation. Instead, it draws functional inspiration from the brain's effective division of labor between plastic learning and stable integration to construct an effective algorithmic. To this end, we design a collaborative framework consisting of three core modules. First, a hippocampus-inspired rapid association module establishes connections between new and historical views, ensuring the model’s plasticity for new knowledge learning. Second, we introduce a Cognitive Forgetting Module (CFM) that incorporates a decay mechanism functionally similar to human active forgetting, optimizing knowledge fusion by dynamically adjusting historical contributions. Finally, we design a prefrontal cortex-inspired Knowledge Consolidation Module (KCM), which leverages the stability of the temporal tensor to refine short-term knowledge into stable long-term memory. By integrating these three functionally complementary modules, EMIMC achieves remarkable knowledge retention capabilities in view-incremental scenarios. Our main contributions are summarized as follows:

\begin{itemize}
\item  EMIMC is the first to introduce the functional principles of the hippocampus-prefrontal cortex collaborative memory model to incremental multi-view clustering, providing a new paradigm to address the Stability-Plasticity Dilemma.
\item We design novel cognitive forgetting and knowledge consolidation modules. The former mitigates catastrophic forgetting by dynamically adjusting historical weights, while the latter utilizes temporal tensors to mathematically instantiate the refinement of short-term knowledge into stable long-term memory.
\item Extensive experiments show that our proposed method outperforms existing incremental clustering methods.
\end{itemize}

\section{Preliminaries}
\label{sec:formatting}

\subsection{Incremental multi-view clustering}  Incremental Multi-View Clustering (IMVC) continuously integrates new views while retaining historical knowledge, thereby achieving stable clustering. Existing literature achieves incremental updates by maintaining historical knowledge carriers ~\cite{CMVC,FDMVC,FCMVC,chen2025multi}. For example, CMVC \cite{CMVC} and CAC \cite{yan2024live} maintain consensus partition matrices to achieve knowledge reuse. FDMVC ~\cite{FDMVC} maintains semantically consistent representations for updates. LAIMVC ~\cite{LAIMVC} maintains consensus anchor graphs and aligns new view anchors through permutation matrices. Consistent with existing research, EMIMC is based on the following assumptions: each view contains complete class information, and class growth scenarios are not considered. Based on this, we focus on drawing functional inspiration from cognitive mechanisms to achieve an efficient capture of multi-view information. 

\subsection{Tensor for knowledge consolidation}
In technical implementation, a core challenge of multi-view clustering is fusing information from diverse views to obtain consensus representations. Recently, representing multi-view data as tensor and leveraging low-rank constraints to capture their underlying high-order correlations has become the mainstream technical paradigm in this field ~\cite{ChenWZC24,guo2022logarithmic,long2025tlrlf4mvc,ji2025anchors}. This paradigm provides powerful mathematical tools for modeling knowledge consolidation. However, minimizing the tensor rank is a nonconvex and NP-hard problem ~\cite{zhong2015nonconvex,PengKCKCC24}. To address this challenge, researchers have proposed nonconvex alternatives, which can more efficiently approximate the tensor rank. Let $\{\mathbf{X}_t\}_{t=1}^V\in\mathbb{R}^{n\times d_t}$ be multi-view data with $V$ views, where $d_t$ is the dimension of the view and $n$ is the sample number. We can optimize the following objective function to achieve our goal.
\begin{equation}
\begin{aligned}
&\min_{\mathbf{A}_t,\mathbf{Z}_t}\sum_{t=1}^V\|\mathbf{X}_t-\mathbf{Z}_t\mathbf{A}_t\|_{2,1}+\lambda\|\mathcal{Z}\|_{low}, s.t.\mathcal{Z}=\Phi(\mathbf{Z}_1,..,\mathbf{Z}_t).
\end{aligned}
\end{equation}
where $\mathbf{A}_t\in\mathbb{R}^{m\times d_t}$ ($m$ is the latent space dimension) is the basic matrix and $\mathbf{Z}_t$ is the representation for $t$-view. Besides, the function $\Phi(\cdot)$ constructs a third-order tensor $\mathcal{Z}\in\mathbb{R}^{n\times m \times V}$ by stacking the representations $\mathbf{Z}_t$ from each view, $\|\cdot\|_{low}$ is the rank constraint term. Here, a commonly used nonconvex substitution function ~\cite{lu2015nonconvex} is introduced, which is defined as:
\newtheorem{Definition}{Definition}
\begin{Definition}
For tensor $\mathcal{Z}\in\mathbb{R}^{n_1\times n_2\times n_3}$, the Alternative Rank Minimization Regularization (ARMR) is defined as:
\begin{equation}
\begin{aligned}
\|\mathcal{Z}\|_{ARMR}=\frac{1}{n_3}\sum_{k=1}^{n_3}\sum_{i=1}^{min(n_1,n_2)}\left(\frac{1-e^{- \mathcal{S}_f^k(i,i)}}{1+e^{-\mathcal{S}_f^k(i,i)}}\right).
\end{aligned}
\end{equation}
\label{ARMR}
\end{Definition}
\noindent With the Fast Fourier Transform (FFT) along the third dimension, we can obtain $\mathcal{Z}_f^k$ by $fft(\mathcal{Z})=\frac{1}{n_3}\sum_{k=1}^{n_3}\mathcal{Z}_f^k$. Then, we perform tensor singular value decomposition(t-SVD) to obtain $\mathcal{Z}_f^k=\mathcal{U}*\mathcal{S}_f^k*\mathcal{V}^T$\cite{kilmer2013third}.
\section{The Proposed method}

\begin{figure*}
    \centering
    \includegraphics[width=0.65\linewidth]{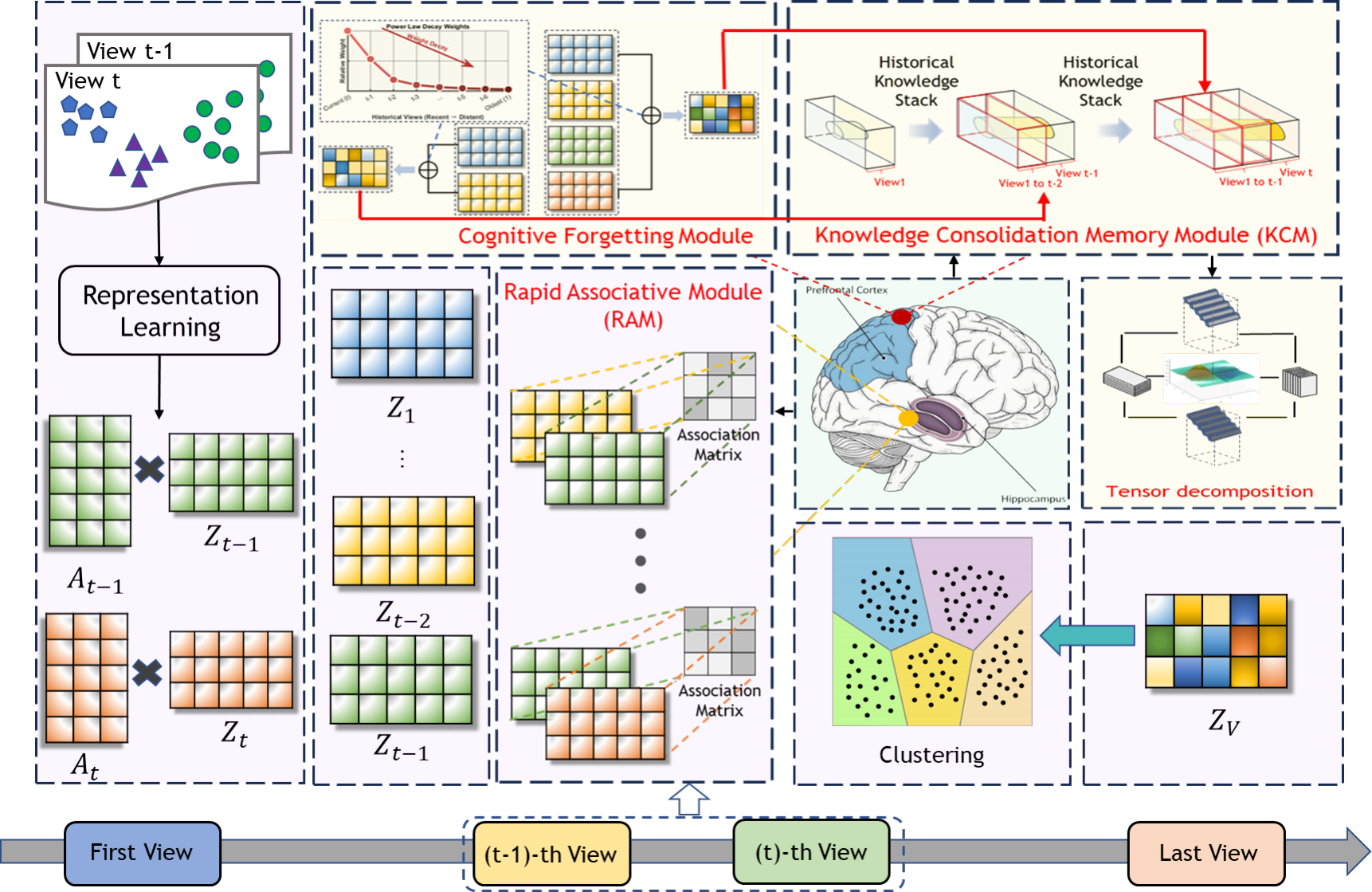}
    \caption{The framework of EMIMC. The model achieves the retention of historical knowledge and the learning of new knowledge through three core modules: Rapid Associative, Cognitive Forgetting, and Knowledge Consolidation Module.}
    \label{fig:enter-label}
\end{figure*}

\subsection{Overview and Motivation}
In incremental multi-view clustering, maintaining a balance between learning new knowledge (plasticity) and retaining existing knowledge (stability) is a major challenge. To address this, we draw functional inspiration from the human brain's collaborative memory mechanism between the hippocampus and the prefrontal cortex. As shown in Figure \ref{fig:enter-label}, EMIMC contains three core components functionally designed to address the Stability-Plasticity Dilemma.
\subsection{Plasticity: New Knowledge Acquisition}
When new view data arrives, the model must quickly adapt to the new knowledge. We functionally emulate the brain's rapid learning ability through incremental representation learning and a hippocampus-inspired rapid associative mechanism.

\noindent \textbf{1) Foundation: Incremental Representation Learning}. First, we design an incremental representation learning strategy. When a new data view $\mathbf{X}_t$ is received, we reconstruct it using a representation $\mathbf{Z}_t$ and an orthogonal basis matrix $\mathbf{A}_t$:
\begin{equation}
\begin{aligned}
\mathcal{L}_{recon}=\|\mathbf{X}_t-\mathbf{Z}_t\mathbf{A}_t\|_{2,1}, \quad s.t. \mathbf{A}_t\mathbf{A}_t^T=\mathbf{I},
\end{aligned}
\end{equation}
where $\mathbf{A}_t\mathbf{A}_t^T=\mathbf{I}$ ensures uncorrelated basis vectors \cite{chen2022efficient}, and the $\ell_{2,1}$ norm enhances robustness \cite{nie2010efficient}.

\noindent \textbf{2) Rapid Associative Module (RAM).}
In cognitive science, a key function of the hippocampus is associative memory, which rapidly establishes connections between newly received information and recent information \cite{mcclelland2020integration}. To simulate this rapid associative capability, we propose a hippocampus-inspired Rapid Associative Module (RAM). This module aims to facilitate immediate interaction between new and old knowledge, enabling the model to adapt to historical contexts while learning new tasks. Specifically, we introduce an orthogonal matrix $\mathbf{P}_t\in\mathbb{R}^{m\times m}$ to connect the current representation $\mathbf{Z}_t$ with the previous representation $\mathbf{Z}_{t-1}$. This process can be expressed as:
\begin{equation}
\begin{aligned}
\mathcal{L}_{associate}=\|\mathbf{Z}_t-{\mathbf{Z}_{t-1}\mathbf{P}_t}\|_F^2, s.t. \mathbf{P}_t\mathbf{P}_t^T = \mathbf{I}.
\end{aligned}
\end{equation}
Although RAM employs a technically mature orthogonal method, this choice is a principled abstraction of the hippocampus. This function simulates the rapid associative processes occurring in the hippocampus: new memories must establish connections with existing memories during the learning process, thereby enabling rapid acquisition of new knowledge.

\subsection{Stability: Knowledge Consolidation and Evolution}
Although plasticity enables systems to adapt to new information, it must be balanced with stability to prevent catastrophic forgetting of accumulated knowledge. To achieve this, we functionally simulate this process through cognitive forgetting and the knowledge consolidation memory module.

\noindent \textbf{1) Cognitive Forgetting Module (CFM).} 

In the human brain, knowledge acquisition is not a simple linear accumulation, but rather involves an active forgetting mechanism. This mechanism continuously removes outdated information to maintain the efficient functioning of the cognitive system. Based on Ebbinghaus's forgetting theory \cite{ebbinghaus1985remembering}, we have introduced a cognitive forgetting module to process view information from past moments. According to this theory, memory decay is initially rapid and then slows down. Therefore, historical knowledge should not contribute equally; instead, its influence should progressively diminish as new views are introduced. Since the mathematical properties of power-law functions exhibit a trend of rapid decline followed by gradual reduction \cite{heathcote2000power}, we utilize this tool to model the decay process. Specifically, for past representation $\mathbf{Z}_i$ (where $i\in\{1,2,...,t-1\}$), we compute the weight $w_i^{(t)}$ using a power-law function based on its relevance to the current view $t$:
\begin{equation}
w_i^{(t)} = \frac{(t-i)^{-\lambda}}{\sum_{j=1}^{t-1} (t-j)^{-\lambda}}, \quad i \in \{1, 2,\dots, t-1\},  
\end{equation}
where $\lambda > 0$ controls the forgetting rate. For view $t$, the integration of historical knowledge can be written as:
\begin{equation}
\mathbf{Z}_{hist} = \sum_{i=1}^{t-1} w_i^{(t)} \mathbf{Z}_i.
\label{Z_hist}
\end{equation}
CFM assigns higher weights to recent representations and decays the contributions of earlier representations. This mechanism functionally mimics the active memory patterns, thereby enhancing historical knowledge expression.

\noindent \textbf{2) Knowledge Consolidation Memory Module (KCM).}

To address long-term stability issues, we draw inspiration from the prefrontal cortex's mechanism of integrating short-term memories into stable patterns. Standard tensor methods typically stack all views into an ever-growing tensor, leading to a cumulative increase in computational complexity. We propose the Knowledge Consolidation Memory (KCM) module. In contrast, our module draws inspiration from the prefrontal cortex's mechanism of integrating interactions between long-term memory and current memory. Specifically, by superimposing the historical representation $\mathbf{Z}_{hist}$ with the current representation $\mathbf{Z}_t$, we construct a temporal tensor $\mathcal{Z}\in\mathbb{R}^{n\times m\times 2}$ to achieve higher-order interaction between old and new knowledge:
\begin{equation}
\begin{aligned}
\mathcal{Z}=stack(\mathbf{Z}_{hist},\mathbf{Z}_{t}, 3),
\end{aligned}
\end{equation}
where $\text{stack}(\cdot, \cdot, 3)$ denotes stacking along the third dimension (where the third dimension has only two slices, among which $\mathcal{Z}(:,:,1) = \mathbf{Z}_{hist}, \mathcal{Z}(:,:,2) = \mathbf{Z}_t$). Next, we need to simulate how the prefrontal cortex consolidates this information into long-term knowledge. To achieve this computationally, we employ the low-rank tensor constraint paradigm discussed in Definition \ref{ARMR}. We treat the tensor $\mathcal{Z}$, formed by stacking historical and current representations, as a high-order knowledge repository. By imposing low-rank constraints on it, we force the model to retain only the most essential structural information across time, functionally simulating the process of knowledge consolidation:
\begin{equation}
\begin{aligned}
\mathcal{L}_{consolidate}=\|\mathcal{Z}\|_{ARMR}, \ s.t. \mathcal{Z}=stack(\mathbf{Z}_{hist},\mathbf{Z}_{t}, 3).
\end{aligned}
\end{equation}

\subsection{The Overall Objective Function}
By combining the functionally inspired modules responsible for plasticity and stability, the overall objective function of EMIMC is formulated as:
\begin{equation}
\begin{aligned}
\mathcal{L}_{EMIMC}=\mathcal{L}_{recon}+\alpha\mathcal{L}_{associate}+\beta\mathcal{L}_{consolidate}.
\end{aligned}
\end{equation}
The final optimization problem is formally stated as:
\begin{equation}
\begin{aligned}
\min_{\substack{\mathbf{Z}_t, \mathbf{A}_t,\\ \mathbf{P}_t, \mathcal{Z}}} &\|\mathbf{X}_t - \mathbf{Z}_t\mathbf{A}_t\|_{2,1} +\alpha\|\mathbf{Z}_t - \mathbf{Z}_{t-1}\mathbf{P}_t\|_F^2+\beta\|\mathcal{Z}\|_{ARMR},\\
 s.t. &\mathbf{Z}_{hist} = \sum_{i=1}^{t-1} \frac{(t-i)^{-\lambda}}{\sum_{j=1}^{t-1} (t-j)^{-\lambda}} \mathbf{Z}_i, \mathbf{A}_t\mathbf{A}_t^T = \mathbf{I},\\
&\mathcal{Z}=stack(\mathbf{Z}_{hist},\mathbf{Z}_t,3),\mathbf{P}_t\mathbf{P}_t^T = \mathbf{I}.
\end{aligned}
\label{eq10}
\end{equation}
\section{Model optimization}
We solve Eq.(\ref{eq10}) using the Alternating Direction Method of Multipliers (ADMM) ~\cite{boyd2011distributed}. It decomposes the original problem into easily solvable subproblems by introducing auxiliary variables and Lagrange multipliers. To facilitate the solution, we introduce auxiliary variables $\mathbf{E}_t$ and $\mathcal{M}$ instead of $\mathbf{X}_t-\mathbf{Z}_t\mathbf{A}_t$ and $\mathcal{Z}$, respectively. Then, the Lagrangian function can be written as
\begin{equation}
\begin{aligned}
&\min_{\mathbf{Z}_t, \mathbf{A}_t, \mathbf{P}_t, \mathcal{Z}} \|\mathbf{E}_t\|_{2,1} + \alpha\|\mathbf{Z}_t - \mathbf{Z}_{t-1}\mathbf{P}_t\|_F^2+\beta\|\mathcal{M}\|_{ARMR}\\
&+\frac{\mu}{2}\|\mathbf{X}_t - \mathbf{Z}_t\mathbf{A}_t-\mathbf{E}_t+\frac{\mathbf{Y}_t}{\mu}\|_F^2+\frac{\rho}{2}\|\mathcal{Z}-\mathcal{M}+\frac{\mathcal{J}}{\rho}\|_F^2,
\end{aligned}
\end{equation}
where $\mathbf{Y}$ and $\mathcal{J}$ denote the Lagrange multiplier, $\mu>0,\rho>0$ are the penalty parameters.
\subsection{Initial view processing (\texorpdfstring{$t = 1$}{t = 1}).}
For the first view, no history information exists, so the objective simplifies to the base view representation learning:\\
\begin{equation}
\begin{aligned}
&\min_{\mathbf{A}_1,\mathbf{E}_1,\mathbf{Z}_1}\frac{\mu}{2}\|\mathbf{X}_1-\mathbf{Z}_1\mathbf{A}_1-\mathbf{E}_1+\frac{\mathbf{Y}_1}{\mu}\|_F^2+\|\mathbf{E}_1\|_{2,1} ,\\
&s.t.\mathbf{A}_1\mathbf{A}_1^T=\mathbf{I}.
\end{aligned}
\label{update_A}
\end{equation}
\textbf{1) Update $\mathbf{A}_1$.} Fixing $\mathbf{Z}_1$, $\mathbf{E}_1$, we can obtain $\mathbf{A}_1$ by: 
\begin{equation}
\begin{aligned}
&\mathop{\arg\min}\limits_{\mathbf{A}_1} \frac{\mu}{2}\|\mathbf{X}_1-\mathbf{Z}_1\mathbf{A}_1-\mathbf{E}_1+\frac{\mathbf{Y}_1}{\mu}\|_F^2 \ s.t. \mathbf{A}_1\mathbf{A}_1^T=\mathbf{I}_m.
\end{aligned}
\label{A1}
\end{equation}
This is a standard Orthogonal Procrustes Problem \cite{wen2018low}. Let $\mathbf{Q}_1 = \mathbf{X}_1 - \mathbf{E}_1 + \frac{\mathbf{Y}_1}{\mu}$. By performing Singular Value Decomposition (SVD) on $\mathbf{Z}_1^T\mathbf{Q}_1$ such that $\mathbf{U}_A\mathbf{S}_A\mathbf{V}_A^T = \mathbf{Z}_1^T\mathbf{Q}_1$, the closed-form solution is given by $\mathbf{A}_1 = \mathbf{U}_A\mathbf{V}_A^T$ (where $\mathbf{U}_A$ and $\mathbf{V}_A$ are the left and right singular matrices).\\
\textbf{2) Update $\mathbf{Z}_1$.} Fixing the other variables, the subproblem of $\mathbf{Z}_1$ can be solved in the following way
\begin{equation}
\begin{aligned}
&\mathop{\arg\min}\limits_{\mathbf{Z}_1} \frac{\mu}{2}\|\mathbf{X}_1-\mathbf{Z}_1\mathbf{A}_1-\mathbf{E}_1+\frac{\mathbf{Y}_1}{\mu}\|_F^2.
\label{Z1}
\end{aligned}
\end{equation}
The problem can be solved in closed form by taking the partial derivatives, and the solution is \\
\begin{equation}
\begin{aligned}
\mathbf{Z}_1=(\mathbf{X}_1-\mathbf{E}_1+\frac{\mathbf{Y}_1}{\mu})\mathbf{A}_1^T.
\end{aligned}
\label{Z2}
\end{equation}

\textbf{3) Update $\mathbf{E}_1$.} Fixing other variables, $\mathbf{E}_1$ can be obtained by 
\begin{equation}
\begin{aligned}
&\mathop{\arg\min}\limits_{\mathbf{E}_1} \frac{\mu}{2}\|\mathbf{X}_1-\mathbf{Z}_1\mathbf{A}_1-\mathbf{E}_1+\frac{\mathbf{Y}_1}{\mu}\|_F^2+\|\mathbf{E}_1\|_{2,1}.\\
\label{E}
\end{aligned}
\end{equation}
The solution of Eq.(\ref{E}) can be obtained by minimizing the thresholding operator in \cite{LTMSC}, which is
\begin{equation}
\begin{aligned}
{(\mathbf{E}_1)}_{(:,i)}=\left\{\begin{matrix}
\frac{\|{(\mathbf{C}_1)}_{(:,i)}\|_{2}-\frac{1}{\mu}} {\|{(\mathbf{C}_1)}_{(:,i)}\|_{2}}{(\mathbf{C}_1)}_{(:,i)},  & if \ \|{(\mathbf{C}_1)}_{(:,i)}\|_{2}>\frac{1}{\mu};\\
 0, & Otherwise.
\end{matrix}\right.
\label{E1}
\end{aligned}
\end{equation}
where $\mathbf{C}_1=\mathbf{X}_1-\mathbf{Z}_1\mathbf{A}_1+\frac{\mathbf{Y}_1}{\mu}$.\\
\textbf{4) Update $\mathbf{Y}_1$ and $\mu$.} The Lagrange multipliers and penalty parameters can be updated by the following rules
\begin{equation}
\begin{aligned}
\mathbf{Y}_1=\mathbf{Y}_1+\mu(\mathbf{X}_1-\mathbf{Z}_1\mathbf{A}_1-\mathbf{E}_1), \mu=min(\delta\mu,\mu_{max}), 
\label{Y}
\end{aligned}
\end{equation}
where $\delta>1$ to accelerate convergence, $\mu_{max}$ is the positive constant. $\mathbf{Z}_1$ will serve as the starting point and first historical memory for the subsequent optimization process.

\subsection{Updates to subsequent views (\texorpdfstring{$t > 1$}{t > 1}).}
\textbf{1) Update $\mathbf{A}_t$ and $\mathbf{E}_t$.} Similarly, the update rules for $\mathbf{A}_t$ and $\mathbf{E}_t$ are similar to the first view from Eq.(\ref{A1}) and Eq.(\ref{E1}). The solution is obtained by $t$ = 2 to $V$ in turn.\\
\textbf{2) Update $\mathbf{P}_t$.} This subproblem can be written as
\begin{equation}
\begin{aligned}
\min_{\mathbf{P}_t} \alpha\|\mathbf{Z}_t-\mathbf{Z}_{t-1}\mathbf{P}_{t}\|_F^2, s.t. \mathbf{P}_t\mathbf{P}_t^T = \mathbf{I}.
\label{P1}
\end{aligned}
\end{equation}
Similar to solving for $\mathbf{A}_t$, we can perform SVD on $\mathbf{C}_t =\mathbf{Z}_{t-1}\mathbf{Z}_{t}^T$ to obtain $\mathbf{P}_t=\mathbf{U}_p\mathbf{V}_p^T$.\\
\textbf{3) Update $\mathcal{M}$.}
When $\mathbf{A}_t, \mathbf{Z}_t, \mathbf{P}_t, \mathbf{E}_t$ are fixed, we can update $\mathcal{M}$ as follow:
\begin{equation}
\begin{aligned}
&\min_{\mathcal{M}}\beta\|\mathcal{M}\|_{ARMR} +\frac{\rho}{2}\|\mathcal{M}-(\mathcal{Z}+\frac{\mathcal{J}}{\rho})\|_F^2.
\label{tensor_Z1}
\end{aligned}
\end{equation}
This is a classic Convex-ConCave Procedure (CCCP) \cite{CCCP}, and we can solve it by the following Lemma:
\newtheorem{lemma}{Lemma}
\begin{lemma}
Let tensor $\mathcal{M}\in\mathbb{R}^{n_1\times n_2 \times n_3}$ with t-SVD. The non-concave tensor rank minimization regularization is solved by:
\begin{equation}
\begin{aligned}
\mathop{\arg\min}\limits_{\mathcal{M}}\frac{\beta}{\rho}\|\mathcal{M}\|_{ARMR}+\frac{1}{2}\|\mathcal{X}-\mathcal{M}\|_F^2.
\label{tensor_Z2}
\end{aligned}
\end{equation}
The optimal solution of Eq.(\ref{tensor_Z2}) can be written as 
\begin{equation}
\begin{aligned}
\mathcal{X}=\Gamma_{\frac{1}{\rho}}(\mathcal{M})=\mathcal{U}*ifft(f(\mathcal{S}_f^k)(i,i))*\mathcal{V}^T,
\label{tensor_Z3}
\end{aligned}
\end{equation}
where \textbf{ifft} denotes the Inverse Fast Fourier Transform. The function $f(\mathcal{S}_f^k)$ is a f-diagonal tensor, which is
\begin{equation}
\begin{aligned}
&f(\mathcal{S}_f^k(i,i),x)=\mathop{\arg\min}\limits_{x\ge 0}\frac{1}{2}(x-\mathcal{S}_f^k(i,i))^2+\left(\frac{1-e^{-x}}{1+e^{- x}}\right).
\label{solve_M_2}
\end{aligned}
\end{equation}
\end{lemma}
Eq. (\ref{solve_M_2}) is a CCCP problem, so we can use the difference of convex programming \cite{tao1997convex} to obtain closed-form solution:
\begin{equation}
\begin{aligned}
f(\mathcal{S}_f^k(i,i))^{iter+1}=(\mathcal{S}_f^k(i,i)-\frac{\nabla  f(\mathcal{S}_f^k(i,i))^{iter}}{\rho})_{+}.
\label{tensor}
\end{aligned}
\end{equation}
where $(\cdot)_{+}$ denotes $max(\cdot,0)$, $\nabla f(\cdot)$ is the gradient of $f(\cdot)$.\\
\textbf{4) Update $\mathbf{Z}_t$.} Fixing the other variables, the optimization function for $\mathbf{Z}_t$ can be written as
\begin{equation}
\begin{aligned}
&\mathop{\arg\min}\limits_{\mathbf{Z}_t} \alpha\|\mathbf{Z}_t - \mathbf{Z}_{t-1}\mathbf{P}_t\|_F^2+\frac{\mu}{2}\|\mathbf{X}_t - \mathbf{Z}_t\mathbf{A}_t-\mathbf{E}_t+\frac{\mathbf{Y}_t}{\mu}\|_F^2\\
&+\frac{\rho}{2}\|\mathbf{Z}_t-\mathbf{M}_t+\frac{\mathbf{J}_t}{\rho}\|_F^2.
\end{aligned}
\end{equation}
This is a convex quadratic function whose optimal solution can be obtained by setting its derivative for $\mathbf{Z}_t$ to zero. Its closed solution can be written as
\begin{equation}
\begin{aligned}
\mathbf{Z}_t&=((\mu+2\alpha+\rho)\mathbf{I})^{-1}(\mu(\mathbf{X}_t-\mathbf{E}_t+\frac{\mathbf{Y}_t}{\mu})\mathbf{A}_t^T\\
&+2\alpha\mathbf{Z}_{t-1}\mathbf{P}_t+\rho\mathbf{M}_t-\mathbf{J}_t).
\label{Z3}
\end{aligned}
\end{equation}\\
\textbf{5) Updating other variables.} The Lagrange multipliers and penalty parameters can be updated by 
\begin{equation}
\begin{aligned}
\mathbf{Y}_t=\mathbf{Y}_t+\mu(\mathbf{X}_t-\mathbf{Z}_t\mathbf{A}_t-\mathbf{E}_t), \mathcal{J}=\mathcal{J}+\rho(\mathcal{Z}-\mathcal{M}).
\label{YY}
\end{aligned}
\end{equation}
\begin{equation}
\begin{aligned}
\mu=min(\delta\mu,\mu_{max}),\rho=min(\delta\rho,\rho_{max}), 
\label{solve_mu}
\end{aligned}
\end{equation}
where $\rho_{max}$ is the positive constant. The solution of our method is summarized in Algorithm \ref{algorithm1}.

\begin{algorithm}[!htbp]
\caption{Updates to subsequent views ($t=2 \to V$)}
\label{algorithm1}
\textbf{Input}: Incremental data $\{\mathbf{X}_t\}_{t=2}^V$, initial representation $\mathbf{Z}_1$, latent dimension $m$, cluster number $k$.\\
\textbf{Parameters}: $\alpha,\beta$, $\lambda$.\\
\textbf{Output}: After obtaining last view $\mathbf{Z}_V$, we use $k$-means to clustering.
\begin{algorithmic}[1]
\FOR{$t = 2$ to $V$}
\STATE \textbf{Initialize}: $\mathbf{A}_t$, $\mathbf{P}_t$, $\mathbf{E}_t$, $\mathbf{Z}_t$, $\mathbf{Y}_t$ = 0. $\mathcal{Z}$, $\mathcal{J}$, $\mathcal{M}$ = 0, $\mu$ = $\rho = 10^{-4}$, $\mu_{max}$ = $\rho_{max}=10^{10}$, $\delta$ = 2.
\WHILE{not converged}
\STATE  Update $\mathbf{Z}_{hist}$ and $\mathbf{A}_t$ by Eq. (\ref{Z_hist}) and Eq. (\ref{A1}).
\STATE  Update $\mathbf{E}_t$ and $\mathbf{P}_t$ by Eq. (\ref{E1}) and Eq. (\ref{P1}).
\STATE  Update $\mathcal{M}$ and $\mathbf{Z}_t$ by Eq. (\ref{tensor}) and Eq. (\ref{Z3}).
\STATE  Update $\mathbf{Y}_t$ and $\mathcal{J}$ by Eq. (\ref{YY}).
\STATE  Update $\mu$ and $\rho$ by Eq. (\ref{solve_mu}).
\ENDWHILE
\ENDFOR
\end{algorithmic}
\end{algorithm}

\subsection{Complexity Analysis}
Our model comprises two stages: the initial view processing ($t = 1$) and the subsequent incremental updates ($t > 1$). For the initial view, the primary computational costs arise from iteratively updating $\mathbf{A}_1$, $\mathbf{Z}_1$, and $\mathbf{E}_1$.
The complexities for these updates are $\mathcal{O}(3dmn+dm^2)$, $\mathcal{O}(dmn)$, $\mathcal{O}(dmn)$, respectively.  Therefore, the complexity for the initial stage is dominated by $\mathcal{O}(dmn+dm^2)$ per iteration. For each subsequent incremental step, the complexities for updating $\mathbf{Z}_{hist}$, $\mathbf{P}_t$, $\mathcal{M}$, $\mathbf{A}_t$, $\mathbf{E}_t$, and $\mathbf{Z}_t$ are $\mathcal{O}(tmn)$, $\mathcal{O}(m^2n+m^3)$,  $\mathcal{O}(m^2n)$, $\mathcal{O}(dmn+dm^2)$, $\mathcal{O}(dmn)$ and 
$\mathcal{O}(dmn+m^2n)$, respectively. Consequently, the overall complexity for an incremental step is $\mathcal{O}(tmn+2dmn+3m^2n+dm^2+m^3)$. Since $d\ll n$ and $m\ll n$, the computational cost grows approximately linearly with $n$. This linear scalability demonstrates the suitability of our model for processing large-scale multi-view datasets.

\section{Experiments}
\noindent \textbf{Datasets and Evaluation.} To assess the efficacy of the proposed method, we conducted experiments on six multi-view datasets, details of which are provided in Table \ref{data}. Three clustering metrics were utilized for performance evaluation: clustering accuracy (ACC), normalized mutual information (NMI), and adjusted rand index (ARI). Higher values indicate better performance.
\begin{table}[!ht]
\caption{The details of Incremental Multi-view datasets.}
\begin{center}
\resizebox{\linewidth}{!}{
\begin{tabular}{c|c|c|c|c}
\hline
Datasets & Samples & Clusters & Views & View Dimensions \\
\hline
ProteinFold & 694 & 12 & 27 & 27, \dots ,27\\
Flower17 & 1360 & 17 & 7 & 1360, \dots ,1360\\
GRAZ02 & 1476 & 4 & 6 & 512,32,256,500,500,680 \\
Handwritten & 2000 & 10 & 6 & 216,76,64,6,240,47\\
Caltech101-all & 9144 & 102 & 5 & 48,40,254,512,928  \\
FMNIST  & 60,000 & 10 & 3 & 1280,512,512 \\
\hline
\end{tabular}}
\end{center}
\label{data}
\end{table}

\begin{table*}[!htbp]
\caption{Performance results for six datasets compared to thirteen state-of-the-art methods ("OM" stands for Out of Memory).}
\resizebox{1\linewidth}{!}{
\centering
\setlength{\tabcolsep}{1mm}
\begin{tabular}{l|ccc|ccc|ccc|ccc|ccc|ccc}
\toprule
\multirow{2}{*}{\textbf{Methods}} & \multicolumn{3}{c|}{\textbf{ProteinFold}} & \multicolumn{3}{c|}{\textbf{Flower17}} & \multicolumn{3}{c|}{\textbf{GRAZ02}} & \multicolumn{3}{c|}{\textbf{Handwritten}} & \multicolumn{3}{c|}{\textbf{Caltech101-all}} & \multicolumn{3}{c}{\textbf{FMNIST}} \\
\cmidrule(lr){2-19} 
~ & \textbf{ACC} & \textbf{NMI} & \textbf{ARI} & \textbf{ACC} & \textbf{NMI} & \textbf{ARI} & \textbf{ACC} & \textbf{NMI} & \textbf{ARI} & \textbf{ACC} & \textbf{NMI} & \textbf{ARI} & \textbf{ACC} & \textbf{NMI} & \textbf{ARI} & \textbf{ACC} & \textbf{NMI} & \textbf{ARI}  \\
\midrule
\multicolumn{19}{c}{Non-Incremental Methods} \\
\midrule
3AMVC & 29.26 & 37.33 & 11.78 & 45.12 & 43.30 & 24.58 & 39.17 & 6.01 & 5.62 & 90.15 & 87.05 & 83.92 & 22.89 & 46.05 & 17.95 & 62.88 & 49.77 & 43.00 \\
NpGC & 37.18 & 43.66 & 17.32 & 58.01 & 57.44 & 40.34 & 42.68 & 8.64 & 8.17 & 89.35 & 80.71 & 78.10 & 28.37 & 49.60 & 24.17 & 63.37 & 46.95 & 41.07 \\
CAMVC & 33.14 & 42.80 & 16.67 & 53.01 & 50.12 & 32.77 & 43.22 & 13.08 & 10.22 & 91.95 & 84.80 & 83.11 & 26.41 & 47.74 & 22.18 & 82.57 & 72.75 & 66.25 \\
IWTSN & 44.09 & 70.59 & 36.27 & 93.27 & 92.93 & 88.79 & 73.71 & 56.68 & 52.96 & 99.65 & 99.10 & 99.22 & 60.94 & 87.55 & 43.85 & 51.58 & 29.45 & 26.12 \\
S$^2$MVTC & 48.27 & 72.45 & 44.48 & 66.03 & 82.53 & 60.41 & 88.96 & 70.60 & 73.14 & 99.90 & 99.76 & 99.78 & 46.50 & 76.22 & 50.89 & 27.74 & 11.88 & 6.57 \\
DSCMC & 29.16 & 39.02 & 12.80 & 34.86 & 44.89 & 23.59 & 38.73 & 5.64 & 5.17 & 71.90 & 68.73 & 58.76 & 24.05 & 42.93 & 18.00 & 61.06 & 40.65 & 37.57 \\
LMTC & 31.63 & 39.50 & 14.29 & 61.38 & 60.97 & 44.09 & 42.35 & 7.71 & 7.36 & 95.10 & 89.23 & 89.40 & 25.20 & 46.53 & 39.08 & 81.70 & 74.72 & 70.00 \\
\midrule
\multicolumn{19}{c}{Incremental Methods} \\
\midrule
CMVC & 31.76 & 40.55 & 14.02 & \underline{55.61} & \underline{55.02} & \underline{38.27} & \underline{48.60} & \underline{12.22} & \underline{12.01} & \underline{89.85} & \underline{80.25} & \underline{78.96} & 24.46 & \underline{46.88} & 20.54 & OM  & OM & OM \\
LAIMVC & 31.89 & 41.32 & 15.27 & 40.37 & 43.83 & 25.56 & 40.88 & 6.95 & 6.42 & 66.98 & 56.46 & 48.40 & 20.34 & 43.40  & 16.42 & 80.30 & \underline{76.81} & 69.75 \\
FDMVC & 36.51 & \underline{47.27} & \underline{19.21} & 47.17 & 50.43 & 31.02 & 43.78 & 9.38 & 8.85 & 79.68 & 77.31 & 69.65 & 21.98 & 44.43 & 18.73 & \underline{81.24} & \textbf{78.22} & \textbf{71.19}\\
FCMVC & 34.01 & 41.96 & 16.12 & 47.22 & 49.73 & 31.47 & 40.40 & 6.34 & 5.93 & 59.59 & 54.57 & 41.54 & \underline{27.05} & 46.87 & \underline{21.62} & OM & OM & OM \\
ASIA & 23.92 & 33.68 & 8.38 & 36.76 & 37.76 & 20.36 & 37.13 & 4.48 & 3.80 & 54.85 & 46.84 & 36.82 & 14.30 & 31.21 & 7.88 & OM & OM & OM  \\
CDIMVC & \underline{37.90} & 44.93 & 17.99 & 40.81 & 45.26 & 23.00 & 41.67 & 9.69 & 8.38 & 79.05 & 72.74 & 53.35 & 20.17 & 44.37 & 13.85 & OM & OM & OM \\
\textbf{Ours} & \textbf{64.06} & \textbf{80.70} & \textbf{54.97} & \textbf{93.37} & \textbf{94.16} & \textbf{89.56} & \textbf{92.28} & \textbf{80.45} & \textbf{80.52} & \textbf{99.79} & \textbf{99.46} & \textbf{99.53} & \textbf{64.46} & \textbf{87.96} & \textbf{52.27} & \textbf{82.86} & 76.37 & \underline{70.80}  \\
\bottomrule
\end{tabular}}
\label{acc1}
\end{table*}

\noindent \textbf{Comparative Methods.} To show the effectiveness of EMIMC, the following methods are selected: 3AMVC \cite{ma2024automatic}(ACM MM'24), NpGC \cite{yu2024non}(AAAI'24), IWTSN \cite{sun2024}(ACM MM'24), S$^2$MVTC\cite{S$^2$MVTC}(CVPR'24), CAMVC \cite{CAMVC}(AAAI'24), DSCMC \cite{kong2025dual}(PR'25), LMTC \cite{liu2025large}(CVPR'25), CMVC \cite{CMVC}(ACM MM'22), LAIMVC \cite{LAIMVC}(ACM MM'24),
FDMVC \cite{FDMVC}(KBS'24),
FCMVC \cite{FCMVC}(TIP'24), ASIA \cite{qu2025anchor}(CSVT'25), CDIMVC \cite{feng2025incremental}(TKDE'25).

\noindent \textbf{Experiment Setup.} All experiments were conducted on a personal computer with an Intel Core i9-13900K CPU and 64 GB of RAM. For all comparison methods, we adopted the parameter settings recommended in their original paper. For our proposed method, we report the average results over five independent runs with different random seeds.

\subsection{Experimental Results and Analysis}
Table \ref{acc1} shows the performance of the proposed method on six datasets with twelve comparison methods. 1) Despite being designed for incremental learning, the proposed method outperforms most traditional non-incremental methods. Notably, the proposed approach achieves performance gains of 26.16\% and 9.94\% in ACC over the second-best method on ProteinFold and Handwritten datasets, respectively. 2) Our method demonstrates improvements over existing incremental methods. These improvements can be attributed to the balance of SPD, thereby enabling efficient knowledge accumulation in incremental tasks.
\subsection{Representation Visualization Analysis}
\begin{figure}[htbp]
  \centering
  \begin{minipage}{0.48\linewidth}
    \centering
    \includegraphics[width=1.0\textwidth]{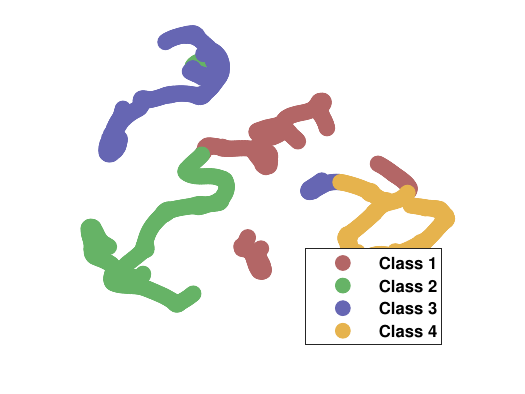}
    \subcaption{IWTSN (ACC = 73.31\%)}
  \end{minipage}
  \begin{minipage}{0.48\linewidth}
    \centering
    \includegraphics[width=1.0\textwidth]{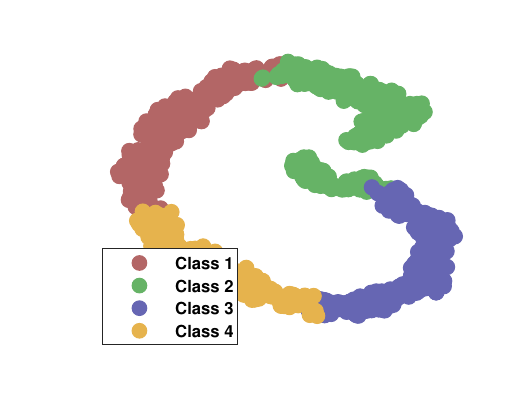}
    \subcaption{Ours (ACC = 92.28\%)}
  \end{minipage}
  \caption{T-SNE visualizations of learned representation.}
  \label{tsne}
\end{figure}

To evaluate the quality of the learned representations, we visualize the GRAZ02 dataset. 
As observed in Figure \ref{tsne}, our method performs better in terms of intra-class cohesion and inter-class dispersion. This is because our collaborative strategy allows the model to retain the historical core knowledge while learning new information, which is crucial for generating high-quality representations.

\subsection{Running Time analysis.} 
To evaluate the computational efficiency of our method, we compare EMIMC with other incremental methods in Table \ref{times} on four datasets. Although LAIMVC excels in speed, EMIMC achieves a trade-off between performance and efficiency. Notably, unlike FCMVC, ASIA and CDIMVC, which fail to complete on large-scale datasets, our method successfully handles challenging cases such as Caltech101-all and FMNIST.
\begin{table}[!htbp]
\caption{Running time (s) comparison on four datasets.}
\centering
\setlength{\tabcolsep}{1mm}
\resizebox{0.9\linewidth}{!}{
\begin{tabular}{l|c|c|c|c}
\toprule
Methods &  \textbf{GRAZ02} & \textbf{Handwritten} & \textbf{Caltech101-all} & \textbf{FMNIST} \\
\midrule
CMVC & 0.44 & 0.63 & 35.56  & OM \\
LAIMVC & \textbf{0.07} & \textbf{0.01} &  \textbf{0.28} & \textbf{2.26} \\
FDMVC & 0.93 & 0.78 & 4.67 & 17.99\\
FCMVC & 1.55 & 1.24  & 86.10 & OM\\
ASIA & 0.69 & 2.99 & 748.76 & OM\\
CDIMVC & 5.81 & 5.10 & 490.21 & OM \\
\textbf{Ours} & \underline{0.22} & \underline{0.17} & \underline{2.00} & \underline{14.30}\\
\bottomrule
\end{tabular}}
\label{times}
\end{table}

\subsection{Incremental
View Analysis.}
\begin{figure}[htbp]
  \centering
  \begin{minipage}{0.48\linewidth}
    \centering
    \includegraphics[width=1.0\textwidth]{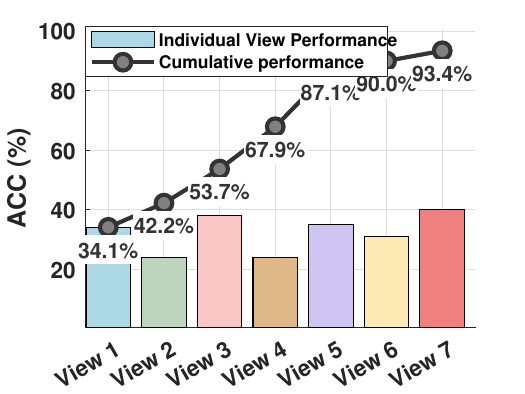}
    \subcaption{Flower17}
  \end{minipage}
  \begin{minipage}{0.48\linewidth}
    \centering
    \includegraphics[width=1.0\textwidth]{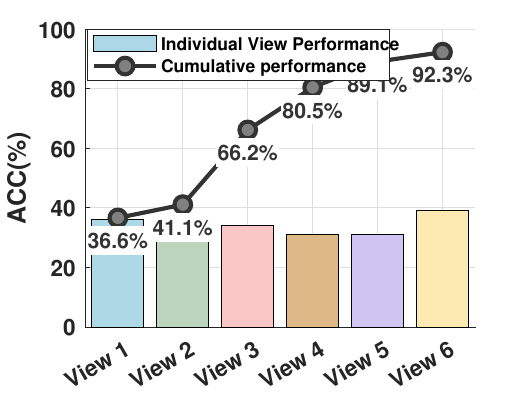}
    \subcaption{GRAZ02}
  \end{minipage}
  \caption{Impact of view number on two datasets.}
  \label{view}
\end{figure}

\noindent \textbf{Impact of view increments.} Figure \ref{view} shows the trend in model performance as the number of views increases, with the single-view results used as a baseline for comparison. As new views are introduced, performance is expected to exhibit a monotonically increasing trend. Experimental results indicate that EMIMC consistently shows gradual improvement as the number of views increases. Notably, our method shows a significant advantage over single-view learning. This phenomenon indicates that our model not only effectively transfers knowledge from existing views but also continuously integrates information from new views.

\begin{figure}[htbp]
  \centering
  \begin{minipage}{0.45\linewidth}
    \centering  \includegraphics[width=1.0\textwidth]{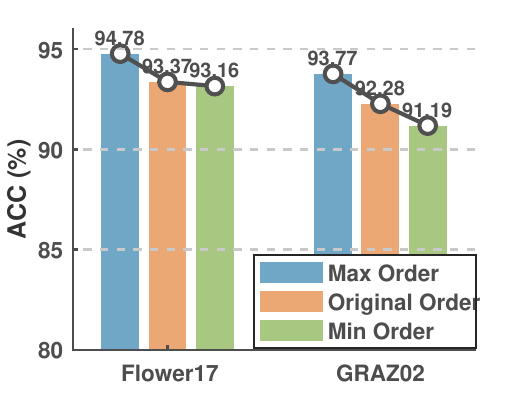}
    \subcaption{Impact of View Order}
    \label{order}
  \end{minipage}
  \begin{minipage}{0.45\linewidth}
    \centering
\includegraphics[width=1.0\textwidth]{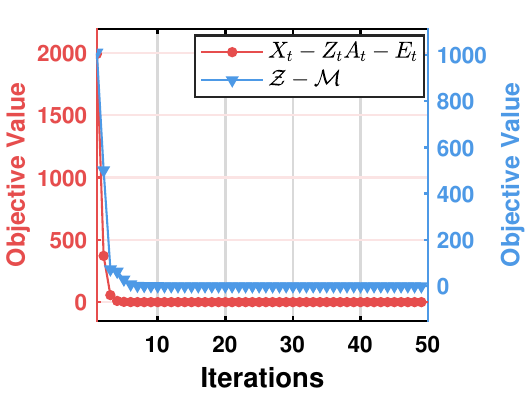} 
    \subcaption{Convergence (FMNIST)}
     \label{obj}
  \end{minipage}
  \caption{The impact of view order and Convergence curve.}
\end{figure}

\noindent \textbf{Impact of view Order.}
As indicated in Figure \ref{view}, significant performance variations exist among individual views. To investigate the impact of view input sequence on performance, we conducted experiments on the Flower17 and GRAZ02 datasets. Given the large number of available views, we evaluated three representative ordering scenarios: Max Order (sorted from best to worst individual performance), Original Order, and Min Order (sorted from worst to best). As shown in Figure \ref{order}, the results indicate that different ordering strategies have a minor impact on performance and consistently outperform the comparison method. This shows the robustness of our method to view order.

\subsection{Convergence Analysis}
To demonstrate convergence, we plot the objective function values against the number of iterations. As observed in Figure \ref{obj}, as the number of iterations increases, the experimental results on the two datasets indicate that the objective function value of the proposed algorithm shows a monotonically decreasing trend as iterations increase and finally converges to a stable point. Thus, the convergence of our method is guaranteed. Due to space constraints, the theoretical proof is shown in the supplementary materials.

\subsection{Ablation Study}
To evaluate the contribution of each module to model performance, we conducted ablation experiments on four datasets in Table \ref{ab}. Our experiments demonstrate the contribution of each module: using only reconstruction loss yields poor performance, confirming its limited discriminative capability. Adding RAM brings modest gains, demonstrating its role in view alignment. KCM drives major improvements, proving its long-term knowledge consolidation capability. Furthermore, CFM also contributes to improvement, suggesting that selectively discarding outdated knowledge can benefit incremental tasks. The collaborative effect of the three modules enables the model to achieve better performance.
\begin{table}[!htbp]
\caption{Ablation experiment results (w/o means without).}
\resizebox{1\linewidth}{!}{
\centering
\setlength{\tabcolsep}{1mm}
\begin{tabular}{l|c|c|c|c}
\toprule
Methods & \textbf{Flower17} & \textbf{GRAZ02} & \textbf{Caltech101-all} & \textbf{FMNIST} \\
\midrule
$\mathcal{L}_{recon}$ & 40.00 & 38.25 & 17.06 & 10.54\\
$\mathcal{L}_{recon}$+RAM  & 42.13 & 38.75 & 17.29 & 10.61\\
$\mathcal{L}_{recon}$+KCM  & 87.06 & 77.37 & 59.22 & 32.13 \\
Ours (w/o CFM)  & 90.74 & 83.20 & 59.28 & 33.97 \\
\textbf{Ours}  & \textbf{93.37}  & \textbf{92.28} & \textbf{64.46} & \textbf{82.86}\\
\bottomrule
\end{tabular}}
\label{ab}
\end{table}

\subsection{Parametric Analysis}
\begin{figure}[htbp]
  \centering
  \begin{minipage}{0.45\linewidth}
    \centering
    \includegraphics[width=1.0\textwidth]{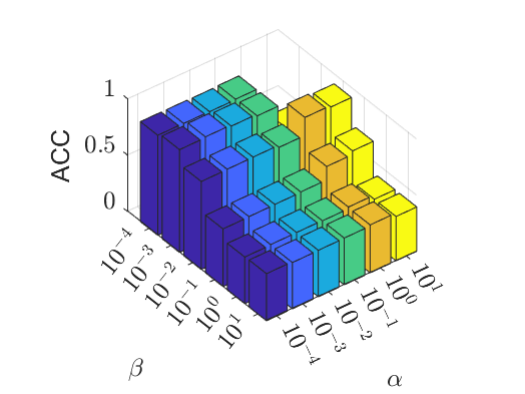}
    \subcaption{GRAZ02}
  \end{minipage}
  \begin{minipage}{0.45\linewidth}
    \centering
    \includegraphics[width=1.0\textwidth]{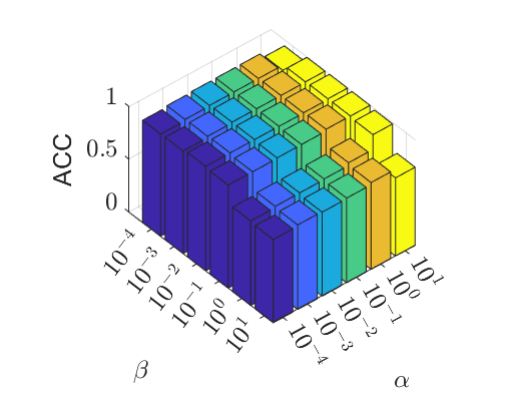}
    \subcaption{Handwritten}
  \end{minipage}
  \caption{Sensitivity analysis of $\alpha$ and $\beta$ on two datasets.}
  \label{alpha}
\end{figure}

\noindent \textbf{Sensitivity analysis of $\alpha$ and $\beta$.} 
To analyze the sensitivity of parameters $\alpha$ and $\beta$, we use the GRAZ02 and Handwritten datasets as representative datasets. With other parameters fixed, we investigate how variations within the interval $\{10^{-4},10^{-3},10^{-2},10^{-1},10^{0},10^{1}\}$ affect the accuracy (ACC) of our method. As shown in Figure \ref{alpha}, our method demonstrates low sensitivity to parameter variations when $\alpha$ and $\beta$ are within $\alpha,\beta \in [10^{-3},10^{-2}]$.

\begin{figure}[htbp]
  \centering
  \begin{minipage}{0.48\linewidth}
    \centering
    \includegraphics[width=1.0\textwidth]{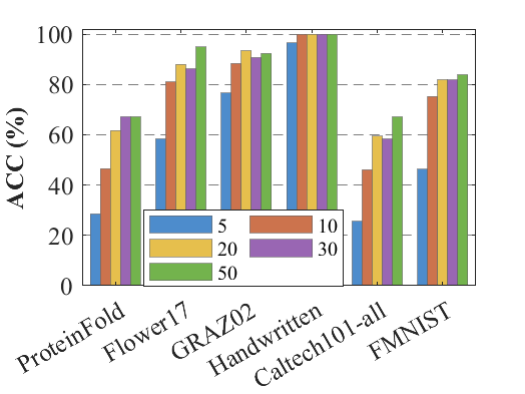}
    \subcaption{Latent space dimension $m$}
    \label{m}
  \end{minipage}
  \begin{minipage}{0.48\linewidth}
    \centering
    \includegraphics[width=1.0\textwidth]{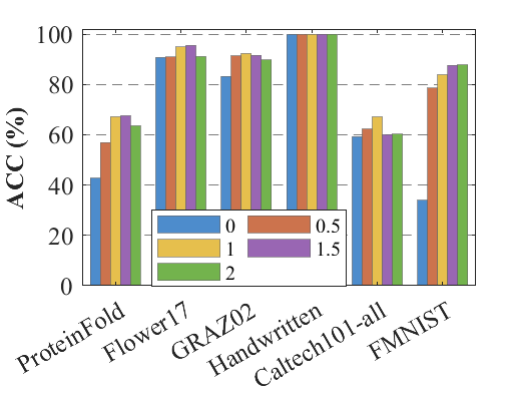}
    \subcaption{Forgetting rate $\lambda$}
    \label{lambda}
  \end{minipage}
  \caption{The impact of $m$ and $\lambda$ on six datasets.}
\end{figure}

\noindent \textbf{Sensitivity analysis of $m$.} 
To investigate the impact of the latent space dimension, we conducted experiments by varying the parameter $m$ in the range of $[5,10,20,30,50]$. From Figure \ref{m}, we can observe that EMIMC exhibits significant robustness to dimensional changes and maintains stable performance across a wide range. In our experiment, we set $m=50$ for all datasets.

\noindent \textbf{Sensitivity analysis of forgetting rate $\lambda$.} To investigate the impact of the forgetting rate $\lambda$, we conducted experiments on the GRAZ02 dataset, comparing different forgetting rate parameters within the range [0, 0.5, 1, 1.5, 2]. As shown in Figure \ref{lambda}, when $\lambda = 0$, the model degenerates to historical representation averaging, yielding the poorest performance due to excessive smoothing of historical representations. Better performance occurs when $\lambda\in[1,1.5]$, indicating that controlled forgetting filters redundant historical information while preserving useful knowledge. In our experiment, we just set $\lambda=1$ for all datasets.

\section{Conclusion}
This paper proposes an incremental multi-view clustering framework based on memory evolution. Our core contribution lies in designing a brain science-inspired component architecture to achieve a balance between plasticity and stability. Compared with current methods, our model demonstrates significant advantages across multiple evaluation metrics, validating its potential for incremental tasks. Although EMIMC performs well in existing incremental scenarios, our model has not explored more challenging scenarios, such as class-incremental learning. In future research, we will develop a class discovery model to eliminate dependence on the number of predefined classes.

{
    \small
    \bibliographystyle{ieeenat_fullname}
    \bibliography{main}
}


\end{document}